# Leveraging Word Embeddings for Spoken Document Summarization

*Kuan-Yu Chen*[*†], *Shih-Hung Liu*[*], *Hsin-Min Wang*[*], *Berlin Chen*[#], *Hsin-Hsi Chen*[†]

[*]Institute of Information Science, Academia Sinica, Taiwan
[#]National Taiwan Normal University, Taiwan
[†]National Taiwan University, Taiwan

[*]{kychen, journey, whm}@iis.sinica.edu.tw, [#]berlin@ntnu.edu.tw, [†]hhchen@csie.ntu.edu.tw

**Abstract**

Owing to the rapidly growing multimedia content available on the Internet, extractive spoken document summarization, with the purpose of automatically selecting a set of representative sentences from a spoken document to concisely express the most important theme of the document, has been an active area of research and experimentation. On the other hand, word embedding has emerged as a newly favorite research subject because of its excellent performance in many natural language processing (NLP)-related tasks. However, as far as we are aware, there are relatively few studies investigating its use in extractive text or speech summarization. A common thread of leveraging word embeddings in the summarization process is to represent the document (or sentence) by averaging the word embeddings of the words occurring in the document (or sentence). Then, intuitively, the cosine similarity measure can be employed to determine the relevance degree between a pair of representations. Beyond the continued efforts made to improve the representation of words, this paper focuses on building novel and efficient ranking models based on the general word embedding methods for extractive speech summarization. Experimental results demonstrate the effectiveness of our proposed methods, compared to existing state-of-the-art methods.

## 1. Introduction

Owing to the popularity of various Internet applications, rapidly growing multimedia content, such as music video, broadcast news programs, and lecture recordings, has been continuously filling our daily life [1-3]. Obviously, speech is one of the most important sources of information about multimedia. By virtue of spoken document summarization (SDS), one can efficiently digest multimedia content by listening to the associated speech summary. Extractive SDS manages to select a set of indicative sentences from a spoken document according to a target summarization ratio and concatenate them together to form a summary [4-7]. The wide spectrum of extractive SDS methods developed so far may be divided into three categories [4, 7]: 1) methods simply based on the sentence position or structure information, 2) methods based on unsupervised sentence ranking, and 3) methods based on supervised sentence classification.

For the first category, the important sentences are selected from some salient parts of a spoken document [8], such as the introductory and/or concluding parts. However, such methods can be only applied to some specific domains with limited document structures. Unsupervised sentence ranking methods attempt to select important sentences based on the statistical features of the sentences or of the words in the sentences without human annotations involved. Popular methods include the vector space model (VSM) [9], the latent semantic analysis (LSA) method [9], the Markov random walk (MRW) method [10], the maximum marginal relevance (MMR) method [11], the sentence significant score method [12], the unigram language model-based (ULM) method [4], the LexRank method [13], the submodularity-based method [14], and the integer linear programming (ILP) method [15]. Statistical features may include the term (word) frequency, linguistic score, recognition confidence measure, and prosodic information. In contrast, supervised sentence classification methods, such as the Gaussian mixture model (GMM) [9], the Bayesian classifier (BC) [16], the support vector machine (SVM) [17], and the conditional random fields (CRFs) [18], usually formulate sentence selection as a binary classification problem, i.e., a sentence can either be included in a summary or not. Interested readers may refer to [4-7] for comprehensive reviews and new insights into the major methods that have been developed and applied with good success to a wide range of text and speech summarization tasks.

Different from the above methods, we explore in this paper various word embedding methods [19-22] for use in extractive SDS, which have recently demonstrated excellent performance in many natural language processing (NLP)-related tasks, such as relational analogy prediction, sentiment analysis, and sentence completion [23-26]. The central idea of these methods is to learn continuously distributed vector representations of words using neural networks, which can probe latent semantic and/or syntactic cues that can in turn be used to induce similarity measures among words, sentences, and documents. A common thread of leveraging word embedding methods to NLP-related tasks is to represent the document (or query and sentence) by averaging the word embeddings corresponding to the words occurring in the document (or query and sentence). Then, intuitively, the cosine similarity measure can be applied to determine the relevance degree between a pair of representations. However, such a framework ignores the inter-dimensional correlation between the two vector representations. To mitigate this deficiency, we further propose a novel use of the triplet learning model to enhance the estimation of the similarity degree between a pair of representations. In addition, since most word embedding methods are founded on a probabilistic objective function, a probabilistic similarity measure might be a more natural choice than non-probabilistic ones. Consequently, we also propose a new language model-based framework, which incorporates the word embedding methods with the document likelihood measure. To recapitulate, beyond the continued and tremendous efforts made to improve the representation of words, this paper focuses on building novel and efficient ranking models on top of the general word embedding methods for extractive SDS.

## 2. Review of Word Embedding Methods

Perhaps one of the most-known seminal studies on developing word embedding methods was presented in [19]. It estimated a statistical (*N*-gram) language model, formalized as a feed-forward neural network, for predicting future words in context while inducing word embeddings (or representations) as a by-product. Such an attempt has already motivated many follow-up extensions to develop similar methods for probing latent semantic and syntactic regularities in the representation of words. Representative methods include, but are not limited to, the continuous bag-of-words (CBOW) model [21], the skip-gram (SG) model [21, 27], and the global vector (GloVe) model [22]. As far as we are aware, there is little work done to contextualize these methods for use in speech summarization.

### 2.1. The Continuous Bag-of-Words (CBOW) Model

Rather than seeking to learn a statistical language model, the CBOW model manages to obtain a dense vector representation (embedding) of each word directly [21]. The structure of CBOW is similar to a feed-forward neural network, with the exception that the non-linear hidden layer in the former is removed. By getting around the heavy computational burden incurred by the non-linear hidden layer, the model can be trained on a large corpus efficiently, while still retains good performance. Formally, given a sequence of words, $w^1, w^2, \ldots, w^T$, the objective function of CBOW is to maximize the log-probability,

$$\sum_{t=1}^{T} \log P(w^t | w^{t-c}, \ldots, w^{t-1}, w^{t+1}, \ldots, w^{t+c}), \quad (1)$$

where $c$ is the window size of context words being considered for the central word $w^t$, $T$ denotes the length of the training corpus, and

$$P(w^t | w^{t-c}, \ldots, w^{t-1}, w^{t+1}, \ldots, w^{t+c}) = \frac{\exp(\mathbf{v}_{\overline{w}^t} \cdot \mathbf{v}_{w^t})}{\sum_{i=1}^{V} \exp(\mathbf{v}_{\overline{w}^t} \cdot \mathbf{v}_{w_i})}, \quad (2)$$

where $\mathbf{v}_{w^t}$ denotes the vector representation of the word $w$ at position $t$; $V$ is the size of the vocabulary; and $\mathbf{v}_{\overline{w}^t}$ denotes the (weighted) average of the vector representations of the context words of $w^t$ [21, 26]. The concept of CBOW is motivated by the distributional hypothesis [27], which states that words with similar meanings often occur in similar contexts, and it is thus suggested to look for $w^t$ whose word representation can capture its context distributions well.

### 2.2. The Skip-gram (SG) Model

In contrast to the CBOW model, the SG model employs an inverse training objective for learning word representations with a simplified feed-forward neural network [21, 28]. Given the word sequence, $w^1, w^2, \ldots, w^T$, the objective function of SG is to maximize the following log-probability,

$$\sum_{t=1}^{T} \sum_{j=-c, j \neq 0}^{c} \log P(w^{t+j} | w^t), \quad (3)$$

where $c$ is the window size of the context words for the central word $w^t$, and the conditional probability is computed by

$$P(w^{t+j} | w^t) = \frac{\exp(\mathbf{v}_{w^{t+j}} \cdot \mathbf{v}_{w^t})}{\sum_{i=1}^{V} \exp(\mathbf{v}_{w_i} \cdot \mathbf{v}_{w^t})}, \quad (4)$$

where $\mathbf{v}_{w^{t+j}}$ and $\mathbf{v}_{w^t}$ denote the word representations of the words at positions $t+j$ and $t$, respectively. In the implementations of CBOW and SG, the hierarchical soft-max algorithm [28, 29] and the negative sampling algorithm [28, 30] can make the training process more efficient and effective.

### 2.3. The Global Vector (GloVe) Model

The GloVe model suggests that an appropriate starting point for word representation learning should be associated with the ratios of co-occurrence probabilities rather than the prediction probabilities [22]. More precisely, GloVe makes use of weighted least squares regression, which aims at learning word representations by preserving the co-occurrence frequencies between each pair of words:

$$\sum_{i=1}^{V} \sum_{j=1}^{V} f(X_{w_i w_j})(\mathbf{v}_{w_i} \cdot \mathbf{v}_{w_j} + b_{w_i} + b_{w_j} - \log X_{w_i w_j})^2, \quad (5)$$

where $X_{w_i w_j}$ denotes the number of times words $w_i$ and $w_j$ co-occur in a pre-defined sliding context window; $f(\cdot)$ is a monotonic smoothing function used to modulate the impact of each pair of words involved in the model training; and $\mathbf{v}_w$ and $b_w$ denote the word representation and the bias term of word $w$, respectively.

### 2.4. Analytic Comparisons

There are several analytic comparisons can be made among the above three word embedding methods. First, they have different model structures and learning strategies. CBOW and SG adopt an on-line learning strategy, i.e., the parameters (word representations) are trained sequentially. Therefore, the order that the training samples are used may change the resulting models dramatically. In contrast, GloVe uses a batch learning strategy, i.e., it accumulates the statistics over the entire training corpus and updates the model parameters at once. Second, it is worthy to note that SG (trained with the negative sampling algorithm) and GloVe have an implicit/explicit relation with the classic weighted matrix factorization approach, while the major difference is that SG and GloVe concentrate on rendering the word-by-word co-occurrence matrix but weighted matrix factorization is usually concerned with decomposing the word-by-document matrix [9, 31, 32].

The observations made above on the relation between word embedding methods and matrix factorization bring us to the notion of leveraging the singular value decomposition (SVD) method as an alternative mechanism to derive the word embeddings in this paper. Given a training text corpus, we have a word-by-word co-occurrence matrix $\mathbf{A}$. Each element $\mathbf{A}_{ij}$ of $\mathbf{A}$ is the log-frequency of times words $w_i$ and $w_j$ co-occur in a pre-defined sliding context window in the corpus. Subsequently, SVD decomposes $\mathbf{A}$ into three sub-matrices:

$$\mathbf{A} \approx \mathbf{U} \Sigma \mathbf{V}^{\mathrm{T}} = \widetilde{\mathbf{A}}, \quad (6)$$

where $\mathbf{U}$ and $\mathbf{V}$ are orthogonal matrices, and $\Sigma$ is a diagonal matrix. Finally, each row vector of matrix $\mathbf{U}$ (or the column vector of matrix $\mathbf{V}^{\mathrm{T}}$, $\mathbf{U}=\mathbf{V}$ since $\mathbf{A}$ is a symmetric matrix) designates the word embedding of a specific word in the vocabulary. It is worthy to note that using SVD to derive the word representations is similar in spirit to latent semantic analysis (LSA) but using the word-word co-occurrence matrix instead of the word-by-document co-occurrence matrix [33].

## 3. Sentence Ranking based on Word Embeddings

### 3.1. The Triplet Learning Model

Inspired by the vector space model (VSM), a straightforward way to leverage the word embedding methods for extractive SDS is to represent a sentence $S_i$ (and a document $D$ to be summarized) by averaging the vector representations of words occurring in the sentence $S_i$ (and the document $D$) [23, 25]:

$$\mathbf{v}_{S_i} = \Sigma_{w \in S_i} \frac{n(w, S_i)}{|S_i|} \mathbf{v}_w. \quad (7)$$

By doing so, the document $D$ and each sentence $S_i$ of $D$ will have a respective fixed-length dense vector representation, and their relevance degree can be evaluated by the cosine similarity measure.

However, such an approach ignores the inter-dimensional correlation between two vector representations. To mitigate the deficiency of the cosine similarity measure, we employ a triplet learning model to enhance the estimation of the similarity degree between a pair of representations [34-36]. Without loss of generality, our goal is to learn a similarity function, $R(\cdot, \cdot)$, which assigns higher similarity scores to summary sentences than to non-summary sentences, i.e.,

$$R(\mathbf{v}_D, \mathbf{v}_{S_i}^+) > R(\mathbf{v}_D, \mathbf{v}_{S_j}^-), \quad (8)$$

where $\mathbf{v}_{S_i}^+$ denotes the sentence representation (in the form of a column vector) for a summary sentence $S_i$, while $\mathbf{v}_{S_j}^-$ is the representation for a non-summary sentence $S_j$. The parametric ranking function has a bi-linear form as follows:

$$R(\mathbf{v}_D, \mathbf{v}_S) \equiv \mathbf{v}_D^T \mathbf{W} \mathbf{v}_S \quad (9)$$

where $\mathbf{W} \in \mathbb{R}^{d \times d}$, and $d$ is the dimension of the vector representation. By applying the passive-aggressive learning algorithm presented in [34], we can derive the similarity function $R$ such that all triplets obey

$$R(\mathbf{v}_D, \mathbf{v}_{S_i}^+) > R(\mathbf{v}_D, \mathbf{v}_{S_j}^-) + \delta. \quad (10)$$

That is, not only the similarity function will distinguish summary and non-summary sentences, but also there is a safety margin of $\delta$ between them. With $\delta$, a hinge loss function can be defined as

$$loss(\mathbf{v}_D, \mathbf{v}_{S_i}^+, \mathbf{v}_{S_j}^-) = \max\{0, \delta - R(\mathbf{v}_D, \mathbf{v}_{S_i}^+) + R(\mathbf{v}_D, \mathbf{v}_{S_j}^-)\}. \quad (11)$$

Then, $\mathbf{W}$ can be obtained by applying an efficient sequential learning algorithm iteratively over the triplets [34, 35]. With $\mathbf{W}$, sentences can be ranked in descending order of similarity measure, and the top sentences will be selected and sequenced to form a summary according to a target summarization ratio.

### 3.2. The Document Likelihood Measure

A recent school of thought for extractive SDS is to employ a language modeling (LM) approach for the selection of important sentences. A principal realization is to use a probabilistic generative paradigm for ranking each sentence $S$ of a document $D$, which can be expressed by $P(D|S)$. The simplest way is to estimate a unigram language model (ULM) based on the frequency of each distinct word $w$ occurring in $S$, with the maximum likelihood (ML) criterion [37, 38]:

$$P(w|S) = \frac{n(w, S)}{|S|}, \quad (12)$$

where $n(w,S)$ is the number of times that word $w$ occurs in $S$ and $|S|$ is the length of $S$. Obviously, one major challenge facing the LM approach is how to accurately estimate the model parameters for each sentence.

Stimulated by the document likelihood measure adopted by the ULM method, for the various word representation methods studied in this paper, we can construct a new word-based language model for predicting the occurrence probability of any arbitrary word $w_j$. Taking CBOW as an example, the probability of word $w_j$ given another word $w_i$ can be calculated by

$$P(w_j | w_i) = \frac{\exp(\mathbf{v}_{w_j} \cdot \mathbf{v}_{w_i})}{\sum_{w_l \in V} \exp(\mathbf{v}_{w_l} \cdot \mathbf{v}_{w_i})}. \quad (13)$$

As such, we can linearly combine the associated word-based language models of the words occurring in sentence $S$ to form a composite sentence-specific language model for $S$, and express the document likelihood measure as

$$P(D|S) = \Pi_{w_j \in D} \left[ \sum_{w_i \in S} \alpha_{w_i} \cdot P(w_j | w_i) \right]^{n(w_j, D)}, \quad (14)$$

where the weighting factor $\alpha_{w_i}$ is set to be proportional to the frequency of word $w_i$ occurring in sentence $S$, subject to $\sum_{w_i \in S} \alpha_{w_i} = 1$. The sentences offering higher document likelihoods will be selected and sequenced to form the summary according to a target summarization ratio.

## 4. Experimental Setup

The dataset used in this study is the MATBN broadcast news corpus collected by the Academia Sinica and the Public Television Service Foundation of Taiwan between November 2001 and April 2003 [39]. The corpus has been segmented into separate stories and transcribed manually. Each story contains the speech of one studio anchor, as well as several field reporters and interviewees. A subset of 205 broadcast news documents compiled between November 2001 and August 2002 was reserved for the summarization experiments. We chose 20 documents as the test set while the remaining 185 documents as the held-out development set. The reference summaries were generated by ranking the sentences in the manual transcript of a spoken document by importance without assigning a score to each sentence. Each document has three reference summaries annotated by three subjects. For the assessment of summarization performance, we adopted the widely-used ROUGE metrics [40]. All the experimental results reported hereafter are obtained by calculating the F-scores [17] of these ROUGE metrics. The summarization ratio was set to 10%. A corpus of 14,000 text news documents, compiled during the same period as the broadcast news documents, was used to estimate related models compared in this paper, which are CBOW, SG, GloVe, and SVD. A subset of 25-hour speech data from MATBN compiled from November 2001 to December 2002 was used to bootstrap the acoustic training with the minimum phone error rate (MPE) criterion and a training data selection scheme [41]. The vocabulary size is about 72 thousand words. The average word error rate of automatic transcription is about 40%.

## 5. Experimental Results

At the outset, we assess the performance levels of several well-practiced or/and state-of-the-art summarization methods for extractive SDS, which will serve as the baseline systems in this paper, including the LM-based summarization method (i.e., ULM, *c.f.* Eq. (12)), the vector-space methods (i.e., VSM, LSA, and MMR), the graph-based methods (i.e., MRW and LexRank), the submodularity method (SM), and the integer linear programming (ILP) method. The results are illustrated in Table 1, where TD denotes the results obtained based on the manual transcripts of spoken documents and SD denotes the results using the speech recognition transcripts that may contain recognition errors. Several noteworthy observations can be drawn from Table 1. First, the two graph-based methods (i.e., MRW and LexRank) are quite competitive with each other and perform better than the vector-space methods (i.e., VSM, LSA, and MMR) for the TD case. However, for the SD case, the situation is reversed. It reveals that imperfect speech recognition may affect the graph-based methods more than the vector-space methods; a possible reason for such a phenomenon is that the speech recognition errors may lead to inaccurate similarity measures between each pair of sentences. The PageRank-like procedure of the graph-based methods, in turn, will be performed based on these problematic measures, potentially leading to degraded results. Second, LSA, which represents the sentences of a spoken document and the document itself in the latent semantic space instead of the index term (word) space, performs slightly better than VSM in both the TD and SD cases. Third, SM and ILP achieve the best results in the TD case, but only have comparable performance to other methods in the SD case. Finally, ULM shows competitive results compared to other state-of-the-art methods, confirming the applicability of the language modeling approach for speech summarization.

**Table 1.** Summarization results achieved by several well-studied or/and state-of-the-art unsupervised methods.

| Method | Text Documents (TD) | | | Spoken Documents (SD) | | |
|---|---|---|---|---|---|---|
| | ROUGE-1 | ROUGE-2 | ROUGE-L | ROUGE-1 | ROUGE-2 | ROUGE-L |
| ULM | 0.411 | 0.298 | 0.361 | 0.364 | 0.210 | 0.307 |
| VSM | 0.347 | 0.228 | 0.290 | 0.342 | 0.189 | 0.287 |
| LSA | 0.362 | 0.233 | 0.316 | 0.345 | 0.201 | 0.301 |
| MMR | 0.368 | 0.248 | 0.322 | 0.366 | 0.215 | 0.315 |
| MRW | 0.412 | 0.282 | 0.358 | 0.332 | 0.191 | 0.291 |
| LexRank | 0.413 | 0.309 | 0.363 | 0.305 | 0.146 | 0.254 |
| SM | 0.414 | 0.286 | 0.363 | 0.332 | 0.204 | 0.303 |
| ILP | 0.442 | 0.337 | 0.401 | 0.348 | 0.209 | 0.306 |

We now turn to investigate the utilities of three state-of-the-art word embedding methods (i.e., CBOW, SG, and GloVe) and the proposed SVD method (*c.f.* Section 2.4), working in conjunction with the cosine similarity measure for speech summarization. The results are shown in Table 2. From the results, several observations can be made. First, the three state-of-the-art word embedding methods (i.e., CBOW, SG, and GloVe), though with disparate model structures and learning strategies, achieve comparable results to each other in both the TD and SD cases. Although these methods outperform the conventional VSM model, they only achieve almost the same level of performance as LSA and MMR, two improved versions of VSM, and perform worse than MRW, LexRank, SM, and ILP in the TD case. To our surprise, the proposed SVD method outperforms other word embedding methods by a substantial margin in the TD case and slightly in the SD case. It should be noted that the SVD method outperforms not only CBOW, SG, and GloVe, but also LSA and MMR. It even outperforms all the methods compared in Table 1 in the SD case.

In the next set of experiments, we evaluate the capability of the triplet learning model for improving the measurement of similarity when applying word embedding methods in speech summarization. The results are shown in Table 3. From the table, two observations can be drawn. First, it is clear that the triplet learning model outperforms the baseline cosine similarity measure (*c.f.* Table 2) in all cases. This indicates that triplet learning is able to improve the measurement of the similarity degree for sentence ranking and considering the inter-dimensional correlation in the similarity measure between two vector representations is indeed beneficial. Second, "CBOW with triplet learning" outperforms all the methods compared in Table 1 in both the TD and SD cases. However, we have to note that learning **W** in Eq. (9) has to resort to a set of documents with reference summaries; thus the comparison is unfair since all the methods in Table 1 are unsupervised ones. We used the development set (*c.f.* Section 4) to learn **W**. So far, we have not figured out systematic and effective ways to incorporate word embeddings into existing supervised speech summarization methods. We leave this as our future work.

**Table 2.** Summarization results achieved by various word-embedding methods in conjunction with the cosine similarity measure.

| Method | Text Documents (TD) | | | Spoken Documents (SD) | | |
|---|---|---|---|---|---|---|
| | ROUGE-1 | ROUGE-2 | ROUGE-L | ROUGE-1 | ROUGE-2 | ROUGE-L |
| CBOW | 0.369 | 0.224 | 0.308 | 0.365 | 0.206 | 0.313 |
| SG | 0.367 | 0.230 | 0.306 | 0.358 | 0.205 | 0.303 |
| GloVe | 0.367 | 0.231 | 0.308 | 0.364 | 0.214 | 0.312 |
| SVD | 0.409 | 0.265 | 0.342 | 0.374 | 0.215 | 0.319 |

**Table 3.** Summarization results achieved by various word-embedding methods in conjunction with the triplet learning model.

| Method | Text Documents (TD) | | | Spoken Documents (SD) | | |
|---|---|---|---|---|---|---|
| | ROUGE-1 | ROUGE-2 | ROUGE-L | ROUGE-1 | ROUGE-2 | ROUGE-L |
| CBOW | 0.472 | 0.367 | 0.432 | 0.396 | 0.258 | 0.347 |
| SG | 0.404 | 0.284 | 0.348 | 0.374 | 0.223 | 0.321 |
| GloVe | 0.372 | 0.248 | 0.315 | 0.375 | 0.225 | 0.319 |
| SVD | 0.422 | 0.303 | 0.364 | 0.376 | 0.223 | 0.323 |

**Table 4.** Summarization results achieved by various word-embedding methods in conjunction with the document likelihood measure.

| Method | Text Documents (TD) | | | Spoken Documents (SD) | | |
|---|---|---|---|---|---|---|
| | ROUGE-1 | ROUGE-2 | ROUGE-L | ROUGE-1 | ROUGE-2 | ROUGE-L |
| CBOW | 0.456 | 0.342 | 0.398 | 0.385 | 0.237 | 0.333 |
| SG | 0.436 | 0.320 | 0.385 | 0.371 | 0.225 | 0.322 |
| GloVe | 0.422 | 0.309 | 0.372 | 0.380 | 0.239 | 0.332 |
| SVD | 0.411 | 0.298 | 0.361 | 0.364 | 0.222 | 0.313 |

In the last set of experiments, we pair the word embedding methods with the document likelihood measure for extractive SDS. The deduced sentence-based language models were linearly combined with ULM in computing the document likelihood using Eq. (14) [40]. The results are shown in Table 4. Comparing the results to that of the word embedding methods paired with the cosine similarity measure (*c.f.* Table 2), it is evident that the document likelihood measure works pretty well as a vehicle to leverage word embedding methods for speech summarization. We also notice that CBOW outperforms the other three word embedding methods in both the TD and SD cases, just as it had done previously in Table 3 when combined with triplet learning, whereas "SVD with document likelihood measure" does not preserve the superiority as "SVD with triplet learning" (*c.f.* Table 3). Moreover, comparing the results with that of various state-of-the-art methods (*c.f.* Table 1), the word embedding methods with the document likelihood measure are quite competitive in most cases.

## 6. Conclusions & Future Work

In this paper, both the triplet learning model and the document likelihood measure have been proposed to leverage the word embeddings learned by various word embedding methods for speech summarization. In addition, a new SVD-based word embedding method has also been proposed and proven efficient and as effective as existing word embedding methods. Experimental evidence supports that the proposed summarization methods are comparable to several state-of-the-art methods, thereby indicating the potential of the new word embedding-based speech summarization framework. For future work, we will explore other effective ways to enrich the representations of words and integrate extra cues, such as speaker identities or prosodic (emotional) information, into the proposed framework. We are also interested in investigating more robust indexing techniques to represent spoken documents in an elegant way.

# 7. References

[1] S. Furui *et al.*, "Fundamental technologies in modern speech recognition," *IEEE Signal Processing Magazine*, 29(6), pp. 16–17, 2012.

[2] M. Ostendorf, "Speech technology and information access," *IEEE Signal Processing Magazine*, 25(3), pp. 150–152, 2008.

[3] L. S. Lee and B. Chen, "Spoken document understanding and organization," *IEEE Signal Processing Magazine*, vol. 22, no. 5, pp. 42–60, 2005.

[4] Y. Liu and D. Hakkani-Tur, "Speech summarization," *Chapter 13* in *Spoken Language Understanding: Systems for Extracting Semantic Information from Speech*, G. Tur and R. D. Mori (Eds), New York: Wiley, 2011.

[5] G. Penn and X. Zhu, "A critical reassessment of evaluation baselines for speech summarization," in *Proc. of ACL*, pp. 470–478, 2008.

[6] A. Nenkova and K. McKeown, "Automatic summarization," *Foundations and Trends in Information Retrieval*, vol. 5, no. 2–3, pp. 103–233, 2011.

[7] I. Mani and M. T. Maybury (Eds.), *Advances in automatic text summarization*, Cambridge, MA: MIT Press, 1999.

[8] P. B. Baxendale, "Machine-made index for technical literature-an experiment," *IBM Journal*, October, 1958.

[9] Y. Gong and X. Liu, "Generic text summarization using relevance measure and latent semantic analysis," in *Proc. of SIGIR*, pp. 19–25, 2001.

[10] X. Wan and J. Yang, "Multi-document summarization using cluster-based link analysis," in *Proc. of SIGIR*, pp. 299–306, 2008.

[11] J. Carbonell and J. Goldstein, "The use of MMR, diversity based reranking for reordering documents and producing summaries," in *Proc. of SIGIR*, pp. 335–336, 1998.

[12] S. Furui *et al.*, "Speech-to-text and speech-to-speech summarization of spontaneous speech", *IEEE Transactions on Speech and Audio Processing*, vol. 12, no. 4, pp. 401–408, 2004.

[13] G. Erkan and D. R. Radev, "LexRank: Graph-based lexical centrality as salience in text summarization", *Journal of Artificial Intelligent Research*, vol. 22, no. 1, pp. 457–479, 2004.

[14] H. Lin and J. Bilmes, "Multi-document summarization via budgeted maximization of submodular functions," in *Proc. of NAACL HLT*, pp. 912–920, 2010.

[15] K. Riedhammer *et al.*, "Long story short - Global unsupervised models for keyphrase based meeting summarization," *Speech Communication*, vol. 52, no. 10, pp. 801–815, 2010.

[16] J. Kupiec *et al.*, "A trainable document summarizer," in *Proc. of SIGIR*, pp. 68–73, 1995.

[17] J. Zhang and P. Fung, "Speech summarization without lexical features for Mandarin broadcast news", in *Proc. of NAACL HLT, Companion Volume*, pp. 213–216, 2007.

[18] M. Galley, "Skip-chain conditional random field for ranking meeting utterances by importance," in *Proc. of EMNLP*, pp. 364–372, 2006.

[19] Y. Bengio *et al.*, "A neural probabilistic language model," *Journal of Machine Learning Research* (3), pp. 1137–1155, 2003.

[20] A. Mnih and G. Hinton, "Three new graphical models for statistical language modeling," in *Proc. of ICML*, pp. 641–648, 2007.

[21] T. Mikolov *et al.*, "Efficient estimation of word representations in vector space," in *Proc. of ICLR*, pp. 1–12, 2013.

[22] J. Pennington *et al.*, "GloVe: Global vector for word representation," in *Proc. of EMNLP*, pp. 1532–1543, 2014.

[23] D. Tang *et al.*, "Learning sentiment-specific word embedding for twitter sentiment classification" in *Proc. of ACL*, pp. 1555–1565, 2014.

[24] R. Collobert and J. Weston, "A unified architecture for natural language processing: deep neural networks with multitask learning," in *Proc. of ICML*, pp. 160–167, 2008

[25] M. Kageback *et al.*, "Extractive summarization using continuous vector space models," in *Proc. of CVSC*, pp. 31–39, 2014.

[26] L. Qiu *et al.*, "Learning word representation considering proximity and ambiguity," in *Proc. of AAAI*, pp. 1572–1578, 2014.

[27] G. Miller and W. Charles, "Contextual correlates of semantic similarity," *Language and Cognitive Processes*, 6(1), pp. 1–28, 1991.

[28] T. Mikolov *et al.*, "Distributed representations of words and phrases and their compositionality," in *Proc. of ICLR*, pp. 1–9, 2013.

[29] F. Morin and Y. Bengio, "Hierarchical probabilistic neural network language model," in *Proc. of AISTATS*, pp. 246–252, 2005.

[30] A. Mnih and K. Kavukcuoglu, "Learning word embeddings efficiently with noise-contrastive estimation," in *Proc. of NIPS*, pp. 2265–2273, 2013.

[31] O. Levy and Y. Goldberg, "Neural word embedding as implicit matrix factorization," in *Proc. of NIPS*, pp. 2177–2185, 2014.

[32] K. Y. Chen *et al.*, "Weighted matrix factorization for spoken document retrieval," in *Proc. of ICASSP*, pp. 8530–8534, 2013.

[33] M. Afify et al., "Gaussian mixture language models for speech recognition," in *Proc. of ICASSP*, pp. IV-29–IV-32, 2007.

[34] K. Crammer *et al.*, "Online passive-aggressive algorithms," *Journal of Machine Learning Research* (7), pp. 551–585, 2006.

[35] G. Chechik *et al.*, "Large scale online learning of image similarity through ranking," *Journal of Machine Learning Research* (11), pp. 1109–1135, 2010.

[36] M. Norouzi *et al.*, "Hamming distance metric learning," in *Proc. of NIPS*, pp. 1070–1078, 2012.

[37] Y. T. Chen *et al.*, "A probabilistic generative framework for extractive broadcast news speech summarization," *IEEE Transactions on Audio, Speech and Language Processing*, vol. 17, no. 1, pp. 95–106, 2009.

[38] C. Zhai and J. Lafferty, "A study of smoothing methods for language models applied to ad hoc information retrieval," in *Proc. of SIGIR*, pp. 334–342, 2001.

[39] H. M. Wang *et al.*, "MATBN: A Mandarin Chinese broadcast news corpus," *International Journal of Computational Linguistics and Chinese Language Processing*, vol. 10, no. 2, pp. 219–236, 2005.

[40] C. Y. Lin, "ROUGE: Recall-oriented understudy for gisting evaluation." 2003 [Online]. Available: http://haydn.isi.edu/ROUGE/.

[41] G. Heigold *et al.*, "Discriminative training for automatic speech recognition: Modeling, criteria, optimization, implementation, and performance," *IEEE Signal Processing Magazine*, vol. 29, no. 6, pp. 58–69, 2012.